\documentclass[sn-mathphys,Numbered]{sn-jnl}

\usepackage{graphicx}%
\usepackage{multirow}%
\usepackage{amsmath,amssymb,amsfonts}%
\usepackage{amsthm}%
\usepackage{mathrsfs}%
\usepackage[title]{appendix}%
\usepackage{xcolor}%
\usepackage{textcomp}%
\usepackage{manyfoot}%
\usepackage{booktabs}%
\usepackage{algorithm}%
\usepackage{algorithmicx}%
\usepackage{algpseudocode}%
\usepackage{listings}%
\usepackage{lastpage}
\usepackage{makecell}
\usepackage{booktabs}
\usepackage{longtable}
\usepackage{tabularx}
\usepackage{etoolbox}
\usepackage{lmodern}

\begin{document}

\title[Multi-Task Learning for Sparsely-Labeled Time Series: A Case Study on Cold-Hardiness Modeling]{Multi-Task Learning for Sparsely-Labeled Time Series: A Case Study on Cold-Hardiness Modeling}

\author*[1]{\fnm{Aseem} \sur{Saxena}}\email{aseem.bits@gmail.com}

\author[2]{\fnm{Paola} \sur{Pesántez-Cabrera}}\email{p.pesantezcabrera@wsu.edu}

\author[3]{\fnm{Markus} \sur{Keller}}\email{mkeller@wsu.edu}

\author[1]{\fnm{Alan} \sur{Fern}}\email{alan.fern@oregonstate.edu}

\affil*[1]{\orgdiv{School of Electrical Engineering and Computer Science}, \orgname{Oregon State University}, \orgaddress{\city{Corvallis}, \postcode{97331}, \state{OR}, \country{USA}}}

\affil[2]{\orgdiv{School of Electrical Engineering and Computer Science}, \orgname{Washington State University}, \orgaddress{\city{Pullman}, \postcode{99163}, \state{WA}, \country{USA}}}

\affil[3]{\orgdiv{Irrigated Agriculture Research and Extension Center}, \orgname{Washington State University}, \orgaddress{\city{Prosser}, \postcode{99350}, \state{WA}, \country{USA}}}

\abstract{We present a real-world case study of multi-task learning (MTL) for temporal process modeling from limited data with temporally sparse labels. Specifically, we investigate multi-task learning for the important agricultural problem of predicting grape cold hardiness, which is the temperature at which lethal freezing occurs. Cold hardiness changes in response to weather and is difficult to measure directly in the field. Thus, growers rely on predictions to decide when to apply costly frost mitigation measures.  We apply recurrent neural networks (RNNs) for daily cold-hardiness prediction from time series weather data. A major challenge is that the cold hardiness response varies across plant cultivars and ground-truth data for each cultivar is temporally sparse and limited. To address this challenge, we investigate multi-task learning (MTL) approaches for combining data, where different tasks correspond to different cultivars. We develop a variety of MTL architectures and evaluate them in both MTL and transfer learning settings.  Our results show significant differences between architectures and that certain architectures are able to consistently outperform single-task learning and state-of-the-art scientific models. Additionally, we show similar results for the qualitatively different, but related, task of budbreak prediction. Further, improved accuracy for budbreak and cold hardiness is achieved by a single MTL model that simultaneously learns both tasks.}


\maketitle

\section{Introduction}\label{sec1}

In real-world applications such as farm operations management in agriculture, we are often interested in learning temporal processes from sparse data. Data can be sparse both in terms of the number of time series (e.g. seasons) and in terms of the sampling rate of ground truth labels (e.g. bi-weekly field samples). It can thus be difficult to leverage the potential benefits of expressive deep models in these applications due to the typically high data requirements. However, such sparse data sets are often related to critical prediction problems (e.g. from different farms or related crops/cultivars). The main contribution of this work is to present a relevant case study from agriculture, where decision support is critical, to investigate the effectiveness of multi-task learning for such applications.

\begin{figure}[t]
    \centering
    \includegraphics[width=1\columnwidth]{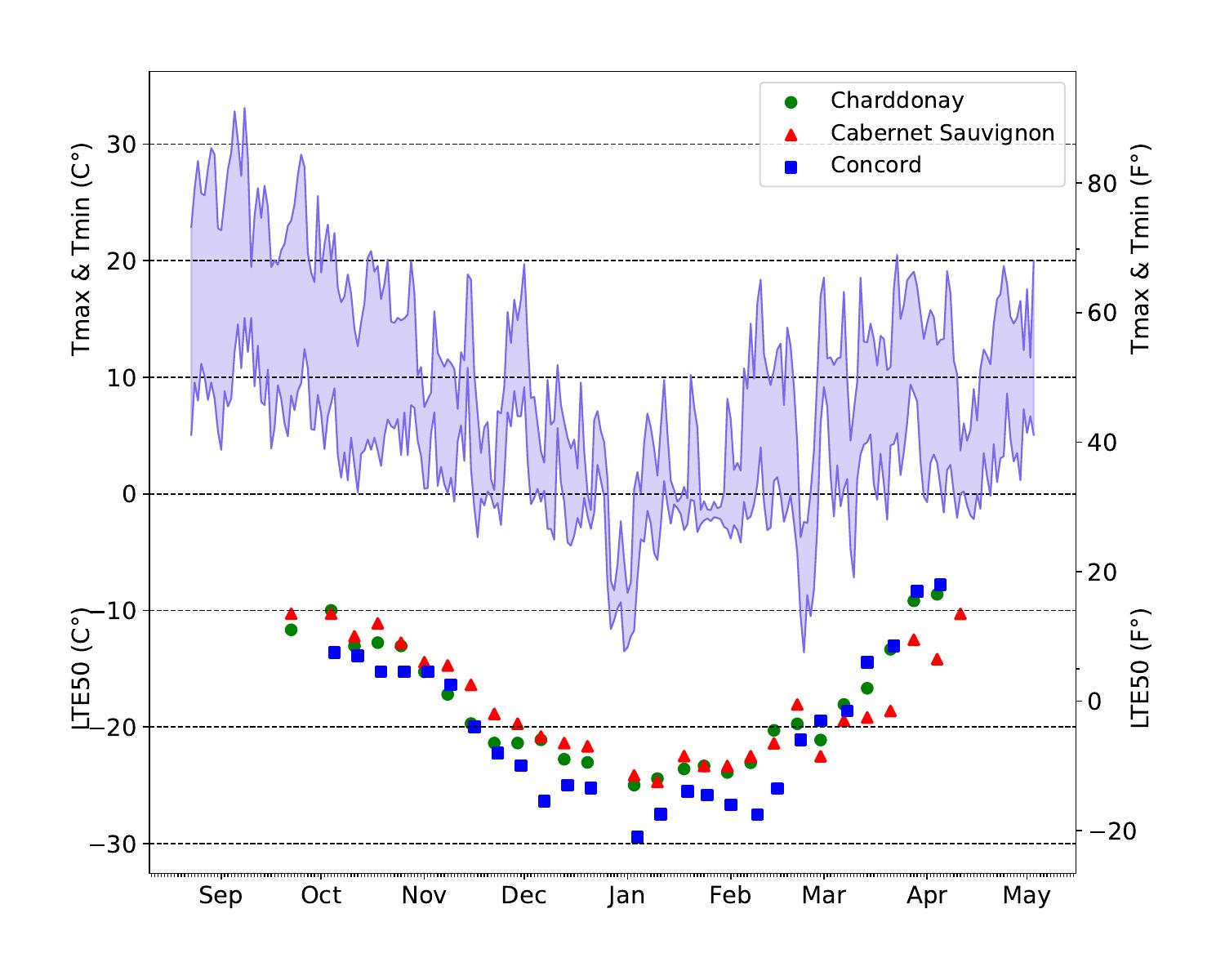} 
    \caption{One season of temperature and cold hardiness measurements at our test vineyard. (Top) Illustrates the daily maximum and minimum temperatures measured at a nearby weather station. (Bottom) $LTE_{50}$ is a cold hardiness measurement that indicates the temperature at which 50\% of the grape buds will freeze. The $LTE_{50}$ measurements are recorded approximately every two weeks and shown for three grape cultivars. The bowl shape indicates an initial increase in cold hardiness (decrease in lethal temperature) during acclimation followed by a decrease in cold hardiness during de-acclimation. Notice the difference in variation of cold hardiness across the different cultivars despite all being subject to the same weather conditions. }
    \label{fig:LTE-example-seasons}
\end{figure}

We investigate the problem of cold hardiness prediction for grapes. Cold hardiness is a measure of the temperature at which lethal freezing occurs in a plant and dictates a plant's ability to survive cold temperatures during the dormant season (September - April). As shown in Figure \ref{fig:LTE-example-seasons}, cold hardiness varies throughout the season, with lethal temperatures decreasing due to acclimation in the fall and increasing due to deacclimation in the early spring leading up to budbreak. Cold hardiness deacclimation rates during low temperatures influence the timing of budbreak. Estimates of cold hardiness and knowledge of the timing of budbreak are essential for growers, as cold hardiness predictions can improve the effectiveness of current frost mitigation measures (i.e., sprinklers, heaters, or wind machines) to avoid cold damage, while knowledge of budbreak helps growers prepare for applications of mitigation to avoid spring frost and reduce pests and disease. Since empirical measurement of cold hardiness requires expensive, specialized equipment, growers rely on models to predict cold tolerance and budbreak. This motivates the need to model cold hardiness and budbreak from historical trends as accurately as possible.

As we overview in Section \ref{sec:science-model}, scientists have developed models of cold hardiness and budbreak using traditional process-style modeling approaches based on biological principles \citep{ferguson_dynamic_2011,ferguson_modeling_2014}. However, the current state-of-the-art is still relatively simple, mostly linear, models and only operates on daily temperature measurements. Rather, cold hardiness responses are also influenced by other weather factors such as humidity and precipitation in non-linear ways. This suggests that there may be benefits to learning more expressive deep models for cold hardiness, which can flexibly incorporate many weather inputs and capture non-linear interactions.

A key challenge for learning deep cold-hardiness models is that the available data is quite limited and temporally sparse. In particular, as shown in Figure \ref{fig:LTE-example-seasons}, cold hardiness and budbreak responses vary across different grape cultivars. While for some cultivars scientists have collected several decades of data, other common cultivars have much less. As our experiments show, deep models trained for ``large data" cultivars can outperform the current state-of-the-art. However, for the ``low data" cultivars, performance can be poor in comparison to current scientific models. One approach to overcoming this low-data problem is to note that while cold-hardiness dynamics vary across cultivars, there is also a common structure since the cultivars are all from a common species. This raises the question of whether we can combine the data across cultivars to improve predictions for each individual cultivar. Importantly this problem structure is present in many other potential applications where there are multiple related processes with limited data and/or sparse labels. Examples include other agricultural applications such as phenology forecasting and soil moisture forecasting, and power system applications such as wind and solar power forecasting \citep{vogt2019wind, schreiber2021emerging, schreiber2021task}.

The main contribution of this paper is a case study on \emph{multi-task learning (MTL)} for leveraging shared structure across cultivars. In particular, we treat different cultivars as different tasks and learn a single model to predict them simultaneously. As reviewed in Section \ref{sec:MTL-Architectures}, a variety of MTL approaches have been proposed with varying levels of implementation and model complexity. To constrain the scope of our case study, we focus on ``wrapper based" MTL approaches that can be easily built on top of an already developed single-task base model. This is in contrast to MTL approaches that either require specific model architectures with many hyperparameters or non-trivial internal modifications to an existing base model. For example, in our application, we put significant work into designing and tuning a recurrent neural network (RNN) for single-cultivar cold-hardiness prediction, which is effective when enough data is available. By focusing on wrapper-based approaches, very little additional work is required to leverage the additional benefits of MTL when data is limited.

We investigate two classes of wrapper-based MTL approaches: \emph{multi-head} and \emph{task embedding}. The multi-head approach interfaces to the base model through the output. It starts with a common base model and attaches task-specific output heads to the model's final feature layer, which can capture a shared task structure. In contrast, task embedding approaches interface to the base model through the input. These approaches learn an embedding vector for each task, which can then be combined with the base model input vector in various ways. In this case, the common structure can be captured via the learned task embedding space. Importantly, each of these MTL approaches can also be easily adapted to a transfer learning scenario, where a new task is introduced after data for the original tasks is unavailable. 

We present experiments on real-world cold hardiness and budbreak data across multiple grape cultivars. We investigate variations of the multi-head and task-embedding approaches and identify configurations that yield consistent improvements compared to single-task models. The improvements are especially pronounced for cultivars with the sparsest data sets. Our results also identify model choices that are consistently less effective. Importantly, the MTL performance is able to consistently outperform the state-of-the-art scientific models, even for low-data cultivars. We also investigate the effectiveness of combining the qualitatively different tasks of budbreak and cold-hardiness prediction via MTL. The results show an additional benefit for both budbreak and cold-hardiness prediction accuracy.  Finally, we show that the same MTL models are effective in the transfer learning setting, allowing for a significant reduction in the amount of data required for new cultivars.

The number of experiments and comparisons is significantly expanded in this paper along with the description of our approaches, including the consideration of MTL across the budbreak and cold-hardiness prediction tasks. In addition, in this paper we extensively study our MTL architecture for use in a transfer learning setting, which demonstrates it as a viable alternative to multi-task learning, when data from prior tasks is not available.

To summarize, the main contributions of this case study are: 1) An empirical investigation of wrapper-based MTL methods on the important real-world problems of cold hardiness and budbreak prediction, which involve small datasets and sparse temporal labeling; 2) Comparing and identifying a subset of the considered MTL approaches that consistently outperform single-task learning; 3) Achieving new state-of-the-art prediction accuracy over the currently used scientific models, noting that one of the MTL models is currently deployed for beta testing by growers on AgWeatherNet\footnote{\url{https://weather.wsu.edu}}, an existing weather network used by growers in the Pacific Northwest; 4) Demonstrating that the most effective MTL models are also highly effective transfer learners, leading to a significant decrease in the required training data.

The paper is organized as follows. Section \ref{sec:science-model} describes the cold hardiness problem and the current prediction approaches. Section \ref{sec:mtlformulation} sets up the notation for MTL with related works and proposed models being discussed. Section \ref{sec:models-and-training} formalizes the cold hardiness and budbreak prediction problem. In Section \ref{sec:datasetdescription}, a brief description of the data sets is provided. Section \ref{sec:experiments} presents our empirical evaluation and main results comparing variations of multi-task and single-task models.  
In Section \ref{sec:deployment} we present the plan for testing and deployment. We conclude with Section \ref{sec:conclusions}.

\section{Cold Hardiness Modeling}
\label{sec:science-model}

The \emph{cold hardiness} of a plant characterizes its ability to resist injury during exposure to low temperatures. In this work, we will focus on the cold hardiness of grapes, where injury corresponds to lethal bud freezing, which decreases crop yield. In order to quantify how cold hardiness varies throughout the dormancy period, scientists use differential thermal analysis (DTA) \cite{mills_cold-hardiness_2006}. This involves putting bud samples in a thermoelectric module, which can sense low temperature exotherms (LTEs) resulting from the freezing of individual buds. The module is placed in a controlled refrigerator, which systematically decreases the temperature as LTEs are monitored. The result is a measurement of the lethal temperatures at which 10\%, 50\%, and 90\% of the bud population die/freeze, which are denoted by $LTE_{10}$, $LTE_{50}$, and $LTE_{90}$, respectively. Figure \ref{fig:LTE-example-seasons} shows the $LTE_{50}$ measurements for three cultivars and temperature ranges throughout a dormant season. The data used in this case study is based on bi-weekly LTE measurements during the dormancy period.

Since this measurement process requires expensive, specialized equipment and expertise, scientists have used collected data to develop grape cold hardiness models, which aim to estimate the lethal temperatures based on only historical temperature data. The current state-of-the-art model, developed by \citet{ferguson_dynamic_2011}, integrates plant biology concepts to find a relation between daily temperatures and changes in cold hardiness. Intuitively, the \emph{Ferguson model} computes the daily change in cold hardiness (e.g. as measured by $LTE_{50}$) based on the day's accumulated thermal time (being above or below certain temperature thresholds) weighted by coefficients that vary with the stage of dormancy. The model has a small number of parameters, e.g. thresholds for thermal times, which can be tuned for a particular cultivar. Tuning was done by performing a brute-force grid search over the parameter space to identify the parameter settings that resulted in the most accurate predictions. While this model has produced promising results and is in use by growers, it has limited expressiveness (only a handful of parameters) and only uses daily temperature data as input, rather than also factoring in other influential weather measurements (e.g., humidity, precipitation, and wind).

In addition to LTE predictions, another important cold-hardiness event is budbreak. During dormancy, the shoot and flower primordia of grapes are protected by bud scales and can reach considerable levels of cold tolerance or hardiness to maximize winter survival \citep{2020keller}. But as spring arrives, the shoots start to grow out during the process of budbreak which results in the emerging green tissues becoming highly vulnerable to frost damage. While it is easy to visually observe budbreak occurrence, the frost vulnerability ushered by it is so sudden that it is detrimental to wait for it to happen without frost measures already in place. Therefore, it is paramount for growers to be able to predict its occurrence ahead of time. This motivates the need for more powerful and accurate budbreak prediction models.

Several scientific models have been proposed for the challenging task of budbreak prediction \citep{nendel_grapevine_2010, ferguson_modeling_2014, zapata_predicting_2017, camargo-a_predicting_2017, leolini_phenological_2020, pina-rey_phenological_2021}. As discussed by \citet{leolini_phenological_2020}, these phenological models can be classified into two main categories: forcing and chilling-forcing models. On one hand, forcing models are based on the accumulation of forcing units from a fixed date in the year. Forcing models focus solely on describing the ecodormancy period by assuming that the endodormancy period has ended and the chilling unit accumulation requirement has been met. On the other hand, chilling-forcing models account for both the endo and ecodormancy periods by considering the chilling unit and the forcing accumulation in relation to specific temperature thresholds. Like the LTE prediction models, these budbreak models only use daily air temperature as an input, rather than also including other potentially relevant environmental factors (e.g., solar radiation, relative humidity, precipitation, dew point). Importantly, different cultivars exhibit different cold hardiness and budbreak responses, which are modeled via a small number of parameters regulating the forcing and chilling models. Available data is used to tune these parameters, typical via exhaustive grid search.

The limitations of the current scientific models for both LTE and budbreak prediction raise the question of whether we can improve cold hardiness and budbreak prediction through the use of modern deep learning models. On one hand, such black-box deep models can be much more expressive and can easily incorporate additional weather data as input. On the other hand, the data set sizes are relatively small from a deep-learning perspective, which may limit the potential benefits, especially when learning cultivar-specific models based on data from single cultivars (i.e. single task learning).

In this work, we address this limited data issue by investigating multi-task learning as an approach to combine data from all cultivars in order to learn better models for individual cultivars. The remainder of the paper exhaustively explores this approach. We turn our attention towards mathematically describing multi-task learning before discussing related works.

\section{Multi-Task Learning}
\label{sec:mtlformulation}
In this section, we first formulate the general problem of multi-task learning in relation to single-task learning. Next, we provide an overview of the primary methods of multi-task learning and motivate the choices of the general methods used for this case study. Later in Section \ref{sec:models-and-training}, we provide details of how the general methods are instantiated for our application domain.

\subsection{Problem Formulation}
\label{Problem Statement}

Traditional machine learning considers \emph{single-task learning (STL)} problems, where the goal is to learn a single predictive mapping from inputs to outputs. For example, in our case study, predicting LTE for a single cultivar from that cultivar's data corresponds to an STL problem. Rather, when multiple related tasks are available, we can consider \emph{multi-task learning (MTL)}, where data can be shared among tasks. For example, in our case study, MTL corresponds to combining LTE data across cultivars to improve overall prediction performance for each individual cultivar. Below, we formalize these notions.

An STL problem is a tuple $(\mathcal{X}, \mathcal{Y}, \mathcal{D}, \mathcal{L})$, where $\mathcal{X}$ is an input space, $\mathcal{Y}$ is an output space, $\mathcal{D} = \{(x_1,y_1), (x_2,y_2), \ldots, (x_N,y_N)\}$ is a data set of input-output pairs sampled from an unknown joint distribution $\mathcal{P}(\mathcal{X},\mathcal{Y})$, and $\mathcal{L}(x,y,y^*)$ is a loss function giving the loss of predicting $y$ for input $x$ when the true prediction target is $y^*$. Given an STL problem, the goal is to return a predictive model $f_{\theta} : \mathcal{X}\rightarrow \mathcal{Y}$ with parameters $\theta$ that achieves small expected loss with respect to $\mathcal{D}$. Typically this is done by attempting to find parameters $\theta^*$ that optimize the empirical training loss,
\[ \theta^* = \arg\min_{\theta} \sum_{(x,y)\in \mathcal{D}} \mathcal{L}(x,f_{\theta}(x),y),\]
possibly combined with a regularization term.


An MTL problem is a collection of STL problems $\{(\mathcal{X}_i, \mathcal{Y}_i, \mathcal{D}_i, \mathcal{L}_i)\}$, where $i$ is the task index. For simplicity of notation, we will assume that the tasks share common input and output spaces $\mathcal{X}$ and $\mathcal{Y}$, i.e. $\mathcal{X}_i = \mathcal{X}$ and $\mathcal{Y}_i = \mathcal{Y}$ for all $i$. Note that each data set $\mathcal{D}_i$ is assumed to be generated by distinct task distributions $\mathcal{P}_i(\mathcal{X},\mathcal{Y})$. For example, for grape LTE prediction each task index corresponds to a single grape cultivar, $\mathcal{X}$ corresponds to varying length weather time series, $\mathcal{Y}$ is the space of real-valued LTE predictions, and $\mathcal{D}_i$ contains input-output pairs for cultivar $i$.

Given an MTL problem, the goal is to return a task-indexed model $f_{\theta}(x, i)$ that returns an output in $\mathcal{Y}$ for an input $x$ with respect to task $i$. The hope is that by allowing the parameters $\theta$ to be informed by all tasks, the predictions produced by $f$ for each individual task will be improved over independent STL solutions for each task. This, of course, requires the tasks to share structure and for the model architecture to be able to capture that structure. The simplest training formulation for MTL is to find parameters $\theta^*$ that optimize the joint training loss across the tasks, i.e.
\[\theta^* = \arg\min_{\theta} \sum_i\sum_{(x,y)\in \mathcal{D}_i} \mathcal{L}_i(x,f_{\theta}(x,i),y)\]
possibly with different weights on the different tasks and additional regularization terms.

Finally, in our experiments, we also consider the \emph{transfer learning} setting for MTL, which is often of practical interest. In transfer learning the tasks are divided into a set of source tasks  $\mathcal{D}_s=\{(\mathcal{X}_i, \mathcal{Y}_i, \mathcal{D}_i, \mathcal{L}_i)\}$ and a target task $(\mathcal{X}', \mathcal{Y}', \mathcal{D}', \mathcal{L}')$. The learning protocol first provides the learning algorithm with the $\mathcal{D}_s$ to produce a model, which we will assume is an MTL model $f_{\theta}(x, i)$. The target task is then made available and a target model $f'(x)$ is produced based on $f_{\theta}$ but without access to the source tasks $\mathcal{D}_s$. A typical example of transfer learning is model fine-tuning in deep learning, where a previously trained model is fine-tuned to predict the labels of a new task. This protocol reflects the scenario where new tasks arrive over time and prior tasks are no longer available, e.g. due to privacy or storage considerations.

\subsection{MTL Methods}
\label{sec:MTL-Architectures}

MTL methods can be broadly partitioned into two classes based on how model parameters are shared among tasks: 1) \emph{Soft parameter sharing}, which does not explicitly share parameters across tasks, and 2) \emph{Hard parameter sharing}, which involves explicit parameter sharing. Below, we overview each class, with an emphasis on hard parameter sharing, which is the focus of this case study.

\subsubsection{Soft Parameter Sharing}

Soft parameter sharing involves starting with an STL model $f_{\theta}$ and then replicating the model for each task via a set of task-specific parameters $\{\theta_i\}$. At test time, predictions for task $i$ are produced via $f_{\theta_i}$. At training time, in order to promote task sharing, the objective function includes terms that encourage the model parameters between tasks to be close to one another. In particular, the MTL loss function typically includes the sum of STL losses for each task in addition to the sum of pairwise distances between task parameters. Various distance metrics have been considered over the parameter space, e.g. the Frobenius Norm \citep{duong2015low} or nuclear norm \citep{yuan2007dimension}.

A potential benefit of soft parameter sharing is that it is highly flexible in allowing individual tasks to specialize. However, this requires identifying the right trade-off between the task-specific losses and the pairwise distance losses. A major drawback of this approach is that it is memory and compute-intensive as the number of tasks grows, especially at training time. For this reason, we do not consider soft parameter sharing in this case study and focus on approaches that incorporate more explicit parameter sharing.

\subsubsection{Hard Parameter Sharing}

Hard parameter sharing is the most widespread paradigm for MTL. Here, a single MTL model architecture is defined, which is able to make predictions for all tasks. The model parameters can then be optimized by simply minimizing the loss across all training data from all tasks. Methods based on hard parameter sharing differ primarily in the choice of model architectures. In general, these methods can be significantly less memory intensive compared to soft parameter sharing, due to explicit sharing of parameters across tasks. A potential downside; however, is that they may be more susceptible to negative interactions among tasks due to sharing. Nevertheless, the practical advantages with respect to computing and memory lead us to focus on hard parameter sharing throughout this study.

In this case study, we are particularly interested in methods based on ``wrapper-style" hard parameter sharing. These methods allow for an MTL architecture to be easily created from an already existing STL architecture for a particular application domain. This has the advantage of allowing for easier MTL adoption in domains where substantial work has already gone into developing an effective STL architecture. Below, we introduce the two wrapper-style approaches considered in this paper, multi-head and task embedding, and then briefly discuss related non-wrapper approaches.

\paragraph{\textbf{Multi-Head Models}}

A multi-head model can be constructed by augmenting a single input and single output neural network model with multiple prediction heads, one per task. Specifically, Figure \ref{fig:model-diagrams}c illustrates a multi-head model where the input is first processed by a shared network $f_{\theta_0}$, which produces a set of shared features. The prediction head for task $i$, denoted $g_{\theta_i}$, then uses those features as input and produces an output for task $i$. Thus, the overall MTL output for an input $x$ and task $i$ is $f(x,i) = g_{\theta_i}(f_{\theta_0}(x))$. The architecture of the shared network $f$ depends on the type of input data, while the prediction head architecture $g$ is typically a shallow Fully Connected Multi Layer Perceptron. Multi-head models are most commonly used for multi-class classification, where each head corresponds to a class label, typically a single linear layer. However, traditional multi-class classification is formulated as an STL problem with a single training set of inputs labeled by their ground-truth classes.

The multi-head model was first proposed for MTL by \cite{caruana1997multitask}. For MTL, each output head corresponds to a distinct task, where the tasks have distinct training sets that may not contain any overlap in their training inputs. \citet{caruana1997multitask} showed that the multi-head model can deliver MTL performance that significantly improves over solving independent STL tasks. The authors hypothesize a number of reasons for the gains, including learning common feature representations that ignore data-dependent noise, which can vary across tasks, and learning to identify the most relevant features as those that are useful across tasks.

A possible drawback of the multi-head model for MTL is that the model size necessarily grows with the number of tasks. This can be particularly problematic when applied in a lifelong learning setting \citep{thrun1995learning} where the number of tasks can grow indefinitely over time. Another potential drawback is that MTL performance can suffer when tasks are not related or only weakly related, which causes the shared network to create a performance bottleneck. Thus, straightforward multi-head models are most appropriate for MTL problems where there is a strong prior belief that tasks are usefully related.

\paragraph{\textbf{Task Embedding Models.}}

Task embedding models for MTL simply augment the input of an STL model with features that specify the task. The most direct instance of this idea is the context-sensitive neural network \citep{silver2008inductive}, which trains an STL neural network $g_{\theta}$ on examples whose inputs are a concatenation $[x,c]$ of the original problem input $x$ and a one-hot binary vector $c$ that indicates the task. That is, if $c_i$ denotes the one-hot vector for $i$, then the MTL model is given by $f_{\theta}(x,i)=g_{\theta}(f_{\theta_0}([x,c_i]))$. Effectively, this approach allows for task-specific biases to be used for network nodes in the input layer, based on the weights associated with $c_i$. This model was shown to improve accuracy over STL models and sometimes outperform multi-head models with equivalent base networks, possibly due to the larger number of parameters required for multiple heads. 

A potential drawback of the context-sensitive neural network is that the one-hot context vector $c$ is maximally sparse and does not reflect any information about task similarity. This limits the degree to which the model can exploit task similarity. One way to address this issue is to use dense task embedding vectors where more related tasks may be assigned closer vectors. The MTL model is given by $f_{\theta}(x,i)=g_{\theta}(f_{\theta_0}([x,E(c_i)]))$, where E is an embedding layer which essentially acts a lookup table to map one-hot $c_i$ vectors to dense vectors $E(c_i)$. This idea, for example, is used in natural language processing, where dense word vectors are used to encode words, with similar words having similar vectors. More related prior work \citep{guo2016entity} studied the use of dense vector embeddings of categorical input features, called entity embeddings. They demonstrated that using entity embeddings, instead of one-hot embeddings, led to better model generalization and lower memory consumption. 

To our knowledge, only three prior works have explored using dense task embeddings for MTL on time series data \citep{vogt2019wind, schreiber2021emerging, schreiber2021task}, where MTL was applied to solar and power estimation problems. In this paper, we extend this idea to consider several variants where the dense task embeddings are learned and combined with the original problem input in various ways. The model first maps a one-hot encoding of the task to a dense embedding vector via a fully connected network layer. This embedding is then combined with the problem input $x$ in one of three ways: concatenation, element-wise addition, or element-wise multiplication. In principle, this architecture allows for task embeddings to be learned end-to-end in a way that best facilitates generalization across tasks.

\paragraph{\textbf{Non-Wrapper Models.}}

From the above description, we see that the multi-head and embedding models are easily instantiated (or wrapped) around a base STL model. Consequently, there have been a number of efforts that attempt to extend these simpler MTL approaches by developing specialized architectures and loss functions. Most of these approaches aim to extend the basic multi-head model \citep{caruana1997multitask} to allow for more flexible types of parameter sharing among tasks. Some examples include cross-stitch networks \citep{misra2016cross}, which allows for limited feature sharing among the task-specific layers, sluice networks \citep{ruder2017sluice} which learns both task-specific and shared feature subspaces across all layers, and the Neural Discriminative Dimensionality Reduction-Convolutional Neural Network (NDDR-CNN) \citep{gao2019nddr}, which generalizes sluice networks via 1-d convolutions.

While these architectures have the potential to improve the accuracy of MTL, the additional complexity yields some important downsides. First, the approaches introduce many additional hyperparameters that can require expertise and experience to tune. Second, the specialized MTL architectures can make it difficult to leverage existing STL architectures developed for particular application domains. For example, our case study involves temporal prediction problems for which we developed a specialized STL architecture (Section \ref{sec:models-and-training}) before the consideration of MTL. This model is not directly compatible with the above non-wrapper models, making it difficult to apply those ideas. In contrast, as described in the next section, it was straightforward to convert our STL model to support MTL via the above wrapper-based methodologies. 

\subsection{Related Work on MTL for Time-Series}

There have been several prior studies of MTL in the context of time-series prediction problems. \citet{mahmoud2020} develop an architecture specialized for time-series classification, where fixed-length time series are classified into a finite number of classes. This approach was applied to activity recognition of human motion data, where the tasks corresponded to different individuals. These models, however, are not easily adaptable to our very different problem of predicting a numeric output times-series based on a growing input time-series. 

MTL has also been applied to time-series forecasting, where the goal is to predict the next $k$ steps of a time-series \cite{YE2019}. Each of the $k$-step predictions was treated as a different task and multi-task learning was used to improve the $k$ step forecasts. Note however, that this work focuses on learning forecasts for a single dynamic process. Thus, this is an orthogonal use of MTL compared to our work, where MTL is applied across multiple processes with distinct, but related, dynamics. 

Recent work \cite{Deng2023} has introduced a multi-view multi-task learning framework for time-series forecasting. In particular, to learn a forecasting model for a process the approach creates multiple artificial sub-tasks based on projecting different views of the data. These sub-tasks are then aggregated via multi-task learning to produce an overall prediction. Here, again, the focus is on forecasting for only a single temporal process and thus the approach is orthogonal to our interest in applying MTL across different processes.

Two recent works have considered MTL for the agricultural time-series problem of predicting wheat yield \cite{Zhuangzhuang2022} and fertilization practices \cite{ZHANG2023}. The motivation for these works is similar to ours, that is, leveraging the relationships between different sources of related agricultural data. However, the approaches are not easily adaptable to our problem. In particular, our time series data contains only sparsely labeled time-series targets, whereas these approaches assume target labels are available for all time-series points. 

As mentioned above the three most closely relate works to ours involve MTL using task embedding models for solar and power estimation \citep{vogt2019wind,schreiber2021emerging, schreiber2021task}. Our work significantly expands on the space of embedding models they consider and provides a direct comparison to multi-head models in addition to considering the transfer setting.

\section{Deep Cold Hardiness and Budbreak Models}
\label{sec:models-and-training}

In this section, we first formalize the cold-hardiness and budbreak prediction problems that are the focus of this case study. Next,  we describe an STL model for this problem based on recurrent neural networks (RNNs), which can be effective when enough training data is available. The MTL instantiations of this model based on the multi-head and task embedding approaches were presented in Section \ref{sec:MTL-Architectures}.

\subsection{Prediction Problems}

Given the availability of cold hardiness, budbreak, and weather data, we can formulate cold-hardiness and budbreak prediction as a sequence prediction problem. We will use $i$ to index the different grape cultivars with $N_i$ denoting the number of seasons available for cultivar $i$. The sequence data for season $k$ of cultivar $i$ is denoted by $S_{i,k}$ and has the form $S_{i,k} = (x_1, e_1, b_1, x_2, e_2, b_2 \ldots, x_H, e_H, b_H)$, where $x_t$ is the weather data for day $t$, $e_t$, $b_t$ are the ground truth LTE and budbreak data for day $t$, and $H$ is the number of days per season. Recall that $e_t$ is not measured on each day of a season (e.g. measured every two weeks) and hence for days where the LTE measurements are unavailable $e_t = N/A$. The budbreak target $b_t$ is 1 if budbreak occurred before or on the day $t$ and is 0 otherwise. Thus, $b_t$ is a step function that rises from 0 to 1 on the day of budbreak. Finally, the data set for cultivar $i$ is denoted by $D_i = \{S_{i,k} \;|\; k\in \{1,\ldots,N_i\}\}$.

In this work, we treat each cultivar as defining a distinct learning task. In particular, the learning task for cultivar $i$ is to produce a model $f_i$ that takes as input a sequence of daily weather measurements $(x_1,x_2,\ldots, x_t)$ up to a particular day $t$ and produces a sequence of predicted LTE estimates $(\hat{e}_1,\hat{e}_2,\ldots,\hat{e}_t)$ and/or a sequence of budbreak occurrence estimates $(\hat{b}_1,\hat{b}_2,\ldots,\hat{b}_t)$ for cultivar $i$. Typically, a farm manager will be most interested in the estimates $\hat{e}_t$ and $\hat{b}_t$ for decision-making. For example, $\hat{e}_t$ can be compared to the low-temperature forecast for that day to help decide whether to prepare for frost mitigation measures. 

A key question motivating this work is whether modern deep learning methods can provide growers and farm managers with improved predictions compared to the current state-of-the-art models.

The STL version of our problem is to learn $f_i$ from only the data in $D_i$. Importantly, the performance of STL is significantly influenced by the amount of available training data, which according to Table \ref{tab:data-description} varies widely across the different cultivars. Indeed, our experiments (Section \ref{sec:experiments}) show that the accuracy of our STL models suffers for cultivars with small data sets. This observation motivated our investigation into MTL where the aim is to learn a predictive model that makes predictions for all cultivars based on the combined cultivar data sets, $\{D_1, D_2, \ldots, D_C\}$, where $C$ is the number of cultivars. Intuitively, since all cultivars derive from a common plant species, it is reasonable to expect that MTL will be able to leverage a common structure in the related prediction problems. Ideally, this will lead to improved performance for individual cultivars, especially those with limited data.

\begin{figure}[t]
    \centering
    \includegraphics[width=1\textwidth]{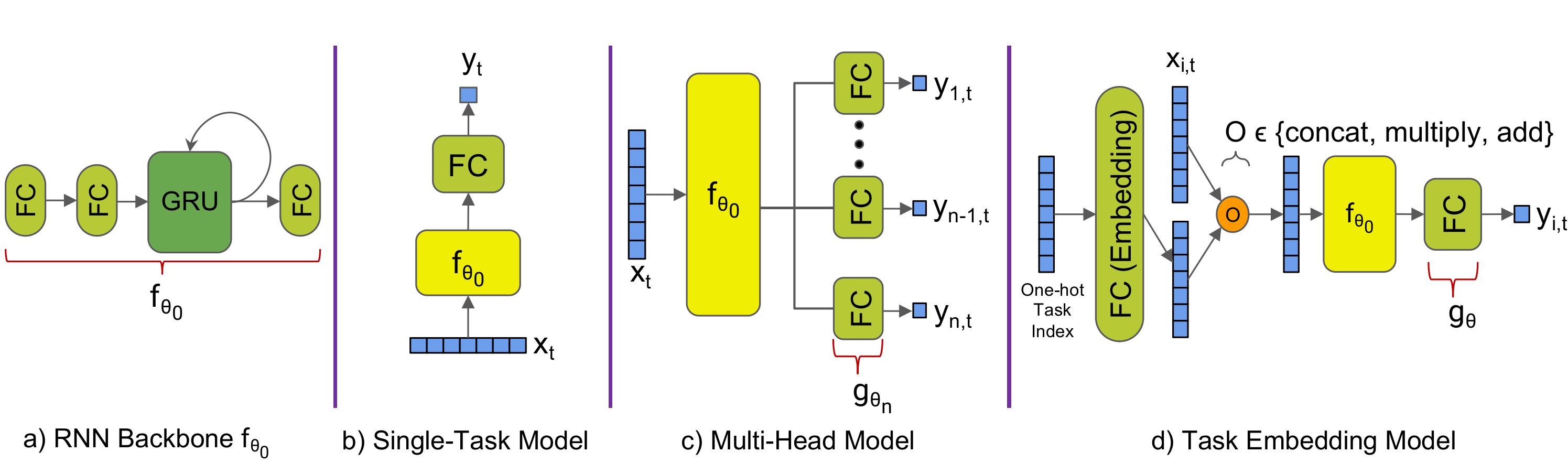} 
    \caption{Network Architectures. FC denotes fully connected layers and GRU denotes Gated Recurrent Unit. a) The RNN backbone is used to process weather data sequences $(x_t)$. b) The single-task model with a single prediction head for a single cultivar. c) Multi-Head Model which has a prediction head for each cultivar allowing backbone features to be shared. d) Task Embedding Model, which combines the weather data features with a learned task embedding for each cultivar before entering the backbone network. Note that the output $y_t$ could indicate cold hardiness $e_t$, budbreak $b_t$, or both. }
    \label{fig:model-diagrams}
\end{figure}

\subsection{Single-Task Model}

Our STL model makes causal LTE and budbreak predictions by sequentially processing a weather data sequence $x_1,x_2,\ldots, x_t$ and at each step outputting the corresponding LTE and/or budbreak estimates.  For this purpose we use a recurrent neural network (RNN) model \cite{RNN}, which is a widely used model for sequence data. The RNN backbone used by both our STL and MTL models is illustrated in Figure \ref{fig:model-diagrams}a, which we denote by $f_{\theta}$ with parameters $\theta$. The backbone network begins with two Fully Connected (FC) layers, followed by a Gated Recurrent Unit (GRU) layer \cite{GRU}, which is followed by another FC layer. Our STL model, shown in Figure \ref{fig:model-diagrams}b, starts with the backbone network $f_{\theta_{0}}$ and then adds an additional fully connected layer, denoted $g_{\theta_{i}}$, that produces a prediction of either the LTE or budbreak. To make a prediction, the daily weather data is fed to the input of $f_{\theta_{0}}$, one day at a time, which produces a feature representation for the sequence $x_1,x_2, \ldots x_t$, denoted $f_{\theta_{0}}(x_1,x_2, \ldots, x_t)$. The final prediction for time $t$ is then given by $g_{\theta_{i}}(f_{\theta_{0}}(x_1, x_2, \ldots, x_t))$.  

Intuitively, the GRU unit, through its recurrent connection is able to build an internal latent-state representation of the sequence data that has been processed so far. For our cold hardiness problem, this representation should capture information about the weather history which is useful for predicting LTE or budbreak. In some sense, the latent state can be thought of as implicitly approximating the internal state of the plant as it evolves during dormancy. As described below, the STL model for cultivar $i$ uses cultivar-specific parameters $\theta_0$ and $\theta_i$, which are trained on that cultivar data set $D_i$.

\subsection{Multi-Task Models for Cold Hardiness and Budbreak}

We consider two types of MTL models that directly extend the RNN backbone of Figure \ref{fig:model-diagrams}a, the multi-head model and the task embedding model.

    {\bf Multi-Head Model.} As illustrated in Figure \ref{fig:model-diagrams}c, the multi-head model is identical to the STL model, except, that it adds $C$ parallel cultivar-specific fully-connected layers to the backbone (i.e. prediction heads). Each prediction head is responsible for producing the LTE or budbreak prediction for its designated cultivar. This model allows the cultivars to share the features produced by the RNN backbone, with each cultivar-specific output simply being a linear combination of the shared features. Intuitively, if there are common underlying features that are useful across cultivars, then this architecture allows those to emerge based on the combined set of data. Thus, cultivars with small amounts of data can leverage those useful features and simply need to tune a set of linear weights based on the available data. We abbreviate this model with \textbf{MultiH} in future sections.

    {\bf Task Embedding Models.} Our proposed Task Embedding models are motivated by thinking about current scientific models and how they address multiple tasks. The Ferguson model\citep{ferguson_dynamic_2011}, for example, has a fixed structure, based on scientific knowledge, but a small number of parameters that can be tuned for each cultivar. As illustrated in Figure \ref{fig:model-diagrams}d, our task embedding model aims to generalize this concept by having a neural network learn both the structure of the model that accepts task-specific parameters as well as learning the parameters of each cultivar. Note that the cultivar parameters and model structure will not have a clear scientific interpretation due to the black-box nature of deep models. The trade-off for lack of interpretability is the potential for better performance due to increased expressive power. This trade-off can be beneficial for the purpose of decision making based on the predictions. In contrast, it is still an open question if and how these models can be used to further scientific understanding and in turn improve mechanistic models.
We explore different ways of incorporating the obtained task embedding, via element-wise multiplication, element-wise addition, and concatenation. We abbreviate these models with \textbf{MultE}, \textbf{AddE}, \textbf{ConcatE} in future sections. In the next section, we describe the data sets we use for applying MTL.

\subsection{Training Details}

We train all of our models using the Adam optimizer \cite{adam} using an initial learning rate of 0.001 with a batch size of 12 seasons shuffled randomly. We train all our models for 400 epochs. The input features have a dimensionality of 12. The output dimensionality of the linear layers of the RNN backbone are 1024, 2048, and 1024, respectively. The GRU has a hidden state and internal memory of dimensionality 2048. Below we present our results. 

\section{Data Set Description}
\label{sec:datasetdescription}
The grapevine data sets for our consideration were collected since the mid '80s generally in the Pacific Northwest region with almost the entirety of the data collection happening in the state of Washington. The data is processed into tabular form with each cultivar having its data in a separate file. We go into further details of the data sets in the subsequent subsections.  

\subsection{Grapes}
\label{subsec:data_grapes}
The cold hardiness and phenological events (including budbreak) of endo\textendash and ecodormant primary buds from up to 30 genetically diverse cultivars/genotypes of field-grown grapevines have been measured since 1988 in the laboratory of the WSU Irrigated Agriculture Research and Extension Center (IAREC) in Prosser, WA (46.29°N latitude; -119.74°W longitude). In the vineyards of the IAREC, the WSU-Roza Research Farm, Prosser, WA (46.25°N latitude, -119.73°W longitude), and in the cultivar collection of Ste. Michelle Wine Estates, Paterson, WA (45.96°N latitude; -119.61°W longitude), cane samples containing dormant buds were collected daily, weekly, or at 2-week intervals from leaf fall in autumn to bud swell in spring. These two phenological events typically occurred in October and in April, respectively \cite{ferguson_dynamic_2011, ferguson_modeling_2014}.

All samples were analyzed with Differential Thermal Analysis (DTA) \citep{mills_cold-hardiness_2006} to record ground truth for $LTE_{10}, LTE_{50}$, and $LTE_{90}$ measurements of cold hardiness. Phenological data were collected as the day of year (DOY) when a particular phenological stage, ranging from bud first swell to harvest, was observed. The budbreak stage is defined as the presence of green tissue in 50\% of previously dormant buds \cite{ferguson_modeling_2014, zapata_predicting_2017}.
Additionally, meteorological/environmental daily data from the closest on-site weather station to each vineyard (cultivar) was obtained using the API provided by AgWeatherNet \cite{AgWeatherNet}. The three stations used are Prosser.NE (46.25°N latitude, -119.74°W longitude), Roza.2 (46.25°N latitude, -119.73°W longitude), and Paterson.E (45.94°N latitude, -119.49°W longitude).

The result is a continually growing data set for each cultivar that contains a varying number of seasons of daily weather data along with cold-hardiness LTE labels for the days that samples were collected along with phenological stage labels placed in the corresponding DOY when observed. Following prior work\citep{ferguson_dynamic_2011, ferguson_modeling_2014}, we consider \emph{a season} to extend from September 7th to May 15th, which is a conservative interval that should almost always contain the full dormancy period. Our experiments involve cultivars with data sets ranging from 34 to 4 seasons.

{\bf Cultivars Data Summary.}
Table \ref{tab:data-description} presents a summary of the number of years of data and the total number of samples collected for the different cultivars. 

\begin{table}[tbp]
    \centering
        \begin{tabular}{ |l|r|r|r| }
            \hline
            \thead{\textbf{Cultivar}}   & \thead{\textbf{LTE Total}}     & \thead{\textbf{Budbreak Total}} & \thead{\textbf{LTE Total}} \\
                               & \thead{\textbf{Years of Data}} & \thead{\textbf{Years of Data}}  & \thead{\textbf{Samples}}   \\\hline
            Barbera            & 14                    & 7                     & 151               \\\hline
            Cabernet Franc     & 4                     & $<$4                      & 35                \\\hline
            Cabernet Sauvignon & 34                    & 21                     & 829               \\\hline
            Chardonnay         & 26                    & 20                     & 783               \\\hline
            Chenin Blanc       & 18                    & 12                     & 193               \\\hline
            Concord            & 27                    & 18                     & 484               \\\hline
            Gewurztraminer     & 9                     & $<$4                      & 101               \\\hline
            Grenache           & 14                    & 7                     & 151               \\\hline
            Lemberger          & 6                     & $<$4                      & 60                \\\hline
            Malbec             & 17                    & 7                     & 261               \\\hline
            Merlot             & 26                    & 21                     & 897               \\\hline
            Mourvedre          & 12                    & 4                     & 133               \\\hline
            Nebbiolo           & 14                    & 7                     & 152               \\\hline
            Pinot Gris         & 17                    & 10                     & 190               \\\hline
            Riesling           & 34                    & 23                     & 636               \\\hline
            Sangiovese         & 15                    & 6                     & 165               \\\hline
            Sauvignon Blanc    & 12                    & 7                     & 140               \\\hline
            Semillon           & 13                    & 7                     & 201               \\\hline
            Syrah              & 23                    & 4                     & 486               \\\hline
            Viognier           & 18                    & 5                     & 206               \\\hline
            Zinfandel          & 14                    & 6                     & 150               \\\hline
        \end{tabular}
    \caption{Summary of LTE and budbreak data collection of grape cultivars.}
    \label{tab:data-description}
\end{table}

{\bf Cultivar Data Set Details.} The data set for a given cultivar contains a row for each day of all data-collection seasons. Note, that since cold hardiness was not measured on each day of a season, some rows do not contain LTE data. Below, we highlight the key information contained in each row used by our models.

\begin{itemize}
    \item AWN\_STATION: The closest AgWeatherNet station from where the environmental readings are taken.
    \item LTE10, LTE50, LTE90 (when available): In degrees Celsius.
    \item MIN\_AT, AVG\_AT, MAX\_AT: Minimum, average, and maximum air temperature observed at 1.5 meters above the ground. In degrees Celsius.
    \item MIN\_RH, AVG\_RH, MAX\_RH: Minimum, average, and maximum relative humidity value observed at 1.5 meters above the ground. In percent.
    \item MIN\_DEWPT, AVG\_DEWPT, MAX\_DEWPT: Minimum, average, and maximum dew point (temperature the air needs to be cooled to in order to achieve relative humidity). In degrees Celsius.
    \item P\_INCHES: Observed sum of precipitation for the daily period. In inches.
    \item WS\_MPH, MAX\_WS\_MPH: Average and maximum observed wind speed at 1.5 meters above the ground for the daily period. In Miles Per Hour.
\end{itemize}

\section{Experiments}
\label{sec:experiments}

In this section, we present in detail the major findings obtained from the experimentation. 
We look at the grapevine data sets for the cold hardiness prediction problem and the budbreak prediction problem. We look at the 21 cultivars in Table \ref{tab:data-description}. Our key finding are summarized below. 

\begin{itemize}
    \item MTL consistently outperforms STL across cultivars and prediction problems.
    \begin{itemize}
        \item Including just 2 seasons of a cultivar's data for training in an MTL setting can match the performance of STL using all of the cultivar's data for grape cold hardiness prediction. 
    \end{itemize}
    \item MTL outperforms the state-of-the-art model\citep{ferguson_dynamic_2011} for most cultivars whereas STL lags behind, especially for cultivars with relatively lower amounts of data. 
    \item MTL can effectively leverage task structure, which is evident from the observation that MTL outperforms an STL baseline model with all the cultivars' data aggregated for training. 
    \item Transfer Learning via Finetuning is a viable alternative to MTL and performs on par with MTL. This is important for real-life scenarios where data for the source tasks is unavailable.
    \item Model Selection is a viable alternative to Finetuning for Transfer Learning. 
    \item There is a synergy between the budbreak and LTE problems and the performance for both tasks improves when trained simultaneously in the MTL setting. 
\end{itemize}

The subsequent subsections will expand into the key findings of our experiments. In \ref{subsec:MTL}, we will look at Multi-Task Learning and how it fares against STL, the current state-of-the-art Ferguson model, and other baselines. In \ref{subsec:TL}, we will look at Transfer Learning methods and how they fare against MTL. Through prior experimentation, we have concluded that among the task embedding model variants, the concatenation embedding model (\textbf{ConcatE}) tends to consistently outperform other variants(\textbf{MultE, AddE}). As a consequence, we only consider the \textbf{ConcatE} model for all our experimental evaluations. We refer the readers to Appendix \ref{sec:lab-comp} to glance at the performance of the MTL embedding model variants for the grape LTE prediction problem.

\begin{table}[b]
\centering
\resizebox{0.9\columnwidth}{!}{
\begin{tabular}{|l|c|c|c|c|c|c|}
\hline
\thead{\textbf{Cultivar}} & \thead{\textbf{ConcatE}} & \thead{\textbf{MultiH}} & \thead{\textbf{Single}} & \thead{\textbf{Ferguson}} & \thead{\textbf{Seasons}} \\ \hline
Barbera            & \textbf{1.50}                       & 1.90                          & 4.22                       & 1.78 & 14                        \\ \hline
Cabernet Franc     & 2.36                        & 2.39                          & 4.00                       & \textbf{1.45} & 4                         \\ \hline
Cabernet Sauvignon & \textbf{1.75}                        & 2.28                          & 3.44                       & 1.83 & 33                        \\ \hline
Chardonnay         & 1.47                        & \textbf{1.40}                          & 1.61                       & 1.79 & 26                        \\ \hline
Chenin Blanc       & 1.51                        & \textbf{1.46}                          & 2.48                       & 2.27 & 18                        \\ \hline
Concord            & 2.42                        & \textbf{1.99}                          & 2.61                       & 2.03 & 26                        \\ \hline
Gewurztraminer     & 1.41                        & \textbf{1.21}                          & 2.71                       & 1.84 & 9                        \\ \hline
Grenache           & 1.87                        & \textbf{1.79}                          & 2.86                       & 1.93 & 14                        \\ \hline
Lemberger          & 1.65                        & \textbf{1.49}                          & 3.24                       & 2.22 & 6                        \\ \hline
Malbec             & 1.32                        & \textbf{0.97}                          & 1.71                       & 1.66 & 17                        \\ \hline
Merlot             & \textbf{1.54}                       & \textbf{1.54}                          & 1.66                       & 1.56 & 26                          \\ \hline
Mourvedre          & 1.66                        & \textbf{1.56}                          & 2.25                       & 1.83 & 12                        \\ \hline
Nebbiolo           & 1.58                        & \textbf{1.24}                          & 2.48                       & 1.80 & 14                        \\ \hline
Pinot Gris         & \textbf{1.61}                        & 1.62                          & 2.04                       & 2.03 & 17                        \\ \hline
Riesling           & \textbf{1.48}                        & 1.97                          & 3.63                       & 1.56 & 34                        \\ \hline
Sangiovese         & 1.74                        & \textbf{1.40}                          & 1.85                       & 1.62 & 15                        \\ \hline
Sauvignon Blanc    & 1.44                        & \textbf{1.23}                          & 1.72                       & 1.43 & 12                        \\ \hline
Semillon           & 1.68                        & 1.76                          & 3.59                       & \textbf{1.51} & 13                        \\ \hline
Syrah              & \textbf{1.22}                        & 1.29                          & 1.57                       & 1.25 & 23                        \\ \hline
Viognier           & 1.75                        & 2.29                          & 4.16                       & \textbf{1.36} & 18                        \\ \hline
Zinfandel          & \textbf{1.45}                        & 1.61                          & 2.64                       & 1.90 & 14                        \\ \hline
Mean               & \textbf{1.64}                        & \textbf{1.64}                          & 2.69                       & 1.75 & 17.38                         \\ \hline
Median             & 1.58                        & \textbf{1.56}                          & 2.61                       & 1.79   & 15                      \\ \hline
\end{tabular}%
}
\caption{Comparison of the performance of proposed MTL methods with STL and the current state-of-the-art Ferguson model for grape cultivars. Note that the performance is measured in terms of Root Mean Squared Error. The model with the lowest error is typeset to bold. The last column shows the number of seasons of data for each cultivar.}
\label{tab:mainresults}
\end{table}

\subsection{Experimental Setup}

We construct our data set by selecting the dormant season data from all cultivars. Missing features are filled in by linear interpolation. We discard seasons that have $ <10\% $ valid LTE readings. We only include seasons where at least $ 90\% $ of temperature data is not missing. Missing LTE label readings are not interpolated, instead, the missing LTE labels' losses are masked during the training and evaluation process. We chose two seasons for each grape cultivar as our test set. We run three trials of training for all our experiments with different train/test splits and average the performance over the three trials.

We rely on the following weather features for learning our models: Air Temperature, Relative Humidity, Dew Point, Precipitation, and Wind Speed. Our models output predictions for $LTE_{10}$, $LTE_{50}$, and $LTE_{90}$ which are optimized simultaneously, helping in inductive transfer. We focus exclusively on the model's predictive power for $LTE_{50}$. We consider the Mean Squared Error (MSE) as our \emph{loss function} and treat the Root Mean Squared Error (RMSE) as our \emph{performance metric}. For the case of budbreak, we consider the Binary Cross Entropy Loss as our loss function for training and evaluation. Additionally, we devise a new metric to measure performance on the test set. In particular, we use a simple approach of predicting budbreak starting on the first day when the softmax output is more than 0.5. The difference in day metric is simply the absolute difference between the predicted day of budbreak and the actual day of budbreak.  We have also considered alternative approaches to estimating the day of budbreak based on the model outputs, but none significantly improve on this simple approach. 

\subsection{Multi-Task Learning}
\label{subsec:MTL}
\subsubsection{MTL vs. STL}
\label{subsubsec:mtlvsstl}

Table \ref{tab:mainresults} shows the root mean squared error (RMSE) of the multi-task models and the single-task model for all 21 grape cultivars.  The first observation is that the multi-task models all outperform the corresponding single-task model across all of the cultivars. Second, we see that the two multi-task have nearly identical median and mean performances in aggregate across cultivars.  

Overall the results for LTE prediction for grapes show that multi-task learning is indeed able to identify and exploit common structures among the different cultivars, leading to improved generalization compared to single task models. Further, we see that in aggregate across grapes, the MultiHead model improves by a larger margin.

\begin{table}[tb]
\centering
\resizebox{0.7\columnwidth}{!}{
\begin{tabular}{|l|r|r|r|}
\hline
Cultivar           & \multicolumn{1}{l|}{ConcatE} & \multicolumn{1}{l|}{MultiH} & \multicolumn{1}{l|}{Single} \\ \hline
Barbera            & \textbf{0.10}     & \textbf{0.10}                        & 2.02                        \\ \hline
Cabernet Sauvignon & 0.12                         & 0.14                        & \textbf{0.07}                        \\ \hline
Chardonnay         & \textbf{0.85}                         & 2.14                        & 1.99                        \\ \hline
Chenin Blanc       & \textbf{0.17}                         & 0.28                        & \textbf{0.17}                        \\ \hline
Concord            & \textbf{0.06}                         & \textbf{0.06}                        & 0.07                        \\ \hline
Grenache           & 0.11                         & \textbf{0.10}                        & 1.30                        \\ \hline
Malbec             & \textbf{0.07}                         & 0.12                        & 3.99                        \\ \hline
Merlot             & \textbf{0.10}                         & 0.56                        & 0.38                        \\ \hline
Mourvedre          & 0.10                         & \textbf{0.05}                        & 10.14                       \\ \hline
Nebbiolo           & \textbf{0.06}                         & 0.13                        & 0.99                        \\ \hline
Pinot Gris         & 0.11                         & 0.13                        & \textbf{0.09}                        \\ \hline
Riesling           & \textbf{0.13}                         & 0.17                        & 0.14                        \\ \hline
Sangiovese         & 0.08                         & \textbf{0.05}                        & 18.62                       \\ \hline
Sauvignon Blanc    & \textbf{0.07}                         & 0.08                        & 0.24                        \\ \hline
Semillon           & \textbf{0.04}                         & 0.05                        & 0.18                        \\ \hline
Syrah              & \textbf{0.01}                         & 0.02                        & 3.89                        \\ \hline
Viognier           & \textbf{0.06}                         & \textbf{0.06}                        & 0.45                        \\ \hline
Zinfandel          & 0.12                         & \textbf{0.09}                        & 0.23                        \\ \hline
Mean          & \textbf{0.13}	& 0.24	& 2.50                        \\ \hline
Median          & \textbf{0.10}	& \textbf{0.10}	& 0.41	 \\ \hline
\end{tabular}
}
\caption{Budbreak Binary Cross Entropy Loss for MTL models and the baseline STL model for each cultivar. The model with the lowest loss is typeset to bold.}
\label{tab:bbresults}
\end{table}

Table \ref{tab:bbresults} shows the Binary Cross Entropy (BCE) loss for the MTL and STL models for all the grape cultivars, note that 3 of the 21 cultivars have not been included due to insufficient data. We observe that one of the MTL variants improves over STL in all cultivars except Cabernet Sauvignon and Pinot Gris. For certain cultivars, STL is better than one or both of the MultiH and ConcatE models. For some cultivars, we see a very large improvement of MTL over STL, e.g. Sangiovese and Syrah. It is possible that cultivar-specific hyperparameter tuning could avoid the poor behavior of STL on these cultivars. However, such tuning is highly undesireable from a practical point of view, given limited data. In contrast, the use of MTL allows for avoiding such tuning by leveraging data from other cultivars. For budbreak prediction, we see that the task embedding model has a better mean performance across cultivars compared to MultiHead, which is largely due to MultiHead performing significantly worse on Chardonnay. Rather, their median performances are similar. We see large STL BCE losses for the Sangiovese and Mourvedre cultivars, this is due to the fact that Sangiovese and Mourvedre have only 4 and 2 seasons of effective training data respectively which leads to deterioration in the STL performance. Similar trends are seen for the Grenache, Malbec, Syrah and Barbera cultivars since they have $<=7$ years of data (Table \ref{tab:data-description}).

To get a better understanding of the practical differences between MTL and STL, we now consider using the models to predict the day of budbreak. In particular, we use a simple approach of predicting budbreak starting on the first day when the softmax output is more than 0.5. The threshold of 0.5 was selected after observing qualitatively similar results for other thresholds. We call this the \textbf{difference in day} metric. Table \ref{tab:dodmetric} shows the median absolute error in day prediction along with the number of predictions that fall beyond selected error thresholds. We see that there are many more outlier predictions with large errors for the STL model compared to the MTL model. We see that the medians for MTL models are significantly better than for STL. Further, the MTL ConcatE and MultiH models produce the fewest larger errors of two weeks or more.

\begin{table}[tb]
\centering
\resizebox{0.9\columnwidth}{!}{
\begin{tabular}{|l|c|c|l|l|l|}
\hline
\textbf{Model} & \textbf{Median} & \textbf{\textgreater{}3 days} & \textbf{\textgreater{}1 week} & \textbf{\textgreater{}2 weeks} & \textbf{\textgreater{}1 month} \\ \hline
Single  & 7   & 32 & \textbf{19} & 13 & 25 \\ \hline
ConcatE & 3.5 & 34 & 24 & \textbf{1}  & \textbf{1}  \\ \hline
MultiH  & \textbf{3}   & \textbf{28} & 30 & \textbf{1}  & \textbf{1}  \\ \hline
\end{tabular}%
}
\caption{Difference of days metric for budbreak prediction. The table provides the median difference in days for predictions corresponding to the first day where the budbreak probability was predicted to be 0.5 or higher. The remaining columns indicate the number of seasons where the budbreak prediction was incorrect by the number of specified days for that column.}
\label{tab:dodmetric}
\end{table}
 
\subsubsection{Comparison to State-of-the-Art} 

From Table \ref{tab:mainresults} we see that both MTL models achieve improved mean and median performance across cultivars compared to the state-of-the-art Ferguson model. Further, with the exception of Cabernet Franc, Semillion and Viognier, at least one of the MTL models outperforms the Ferguson model. Specifically, the Ferguson model is outperformed by the embedding model on 14 cultivars and by the MultiH model on 15 cultivars. This is in contrast to the STL model, which is outperformed by the Ferguson model across all cultivars.

\subsubsection{Impact of Task Data Set Size for Grape LTE Prediction.} 

\begin{figure}[tbp]
    \centering
    \includegraphics[width=1\textwidth]{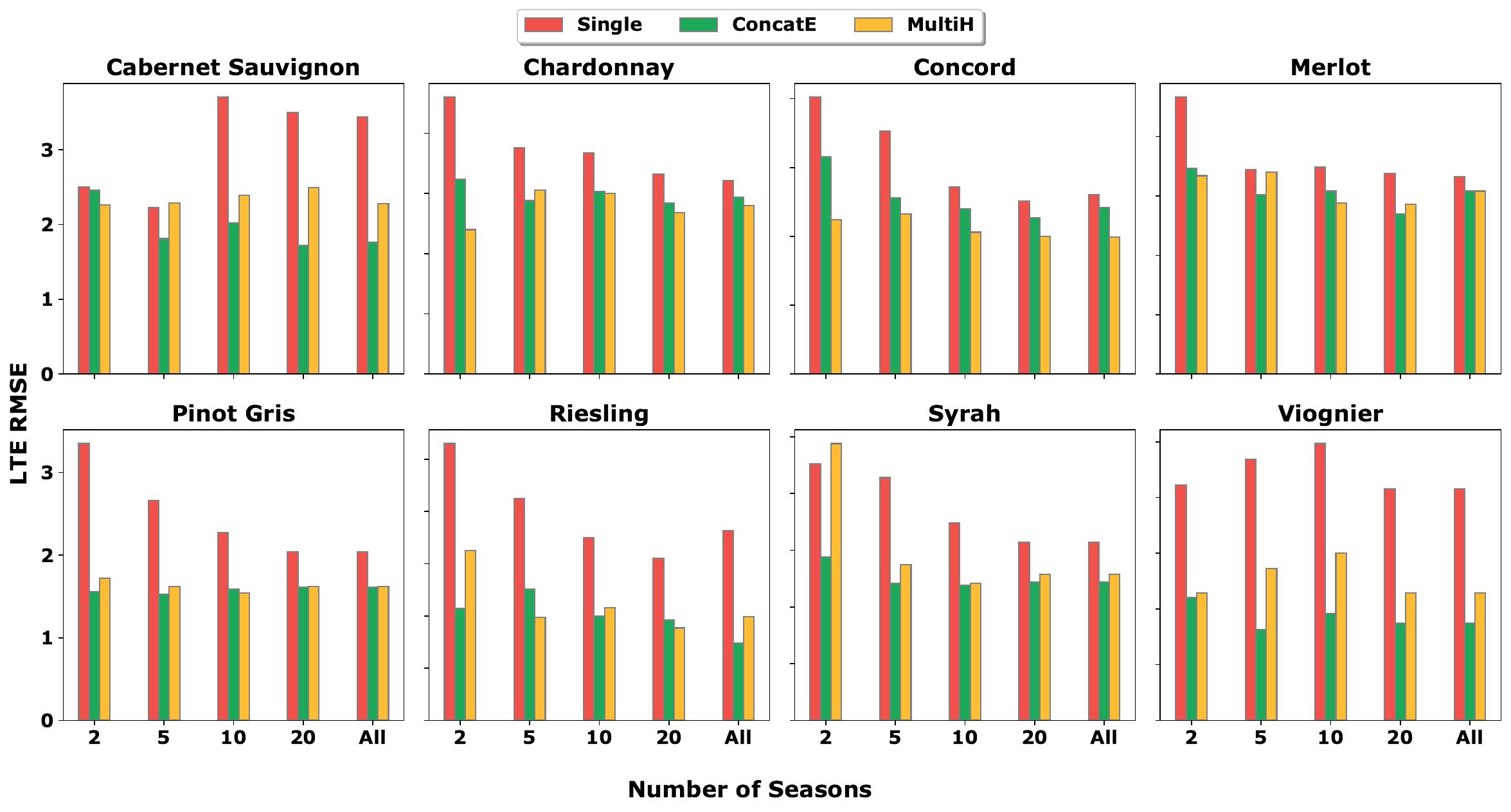} 
    \caption{We look at the cultivars with high amounts of data and artificially restrict the training set size for both the MTL and STL settings. We choose 2, 5, 10, 20, or all of the available seasons in the training set. The x-axis refers to the number of seasons included in the training set and the y-axis denotes the RMSE achieved. The complete set of results is included in Appendix \ref{sec:full-dataset-size}}.
    \label{fig:dataablation}
\end{figure}

In order to observe the impact of amount of training data, we selected 8 cultivars which had the largest number of seasons and trained models using subsets of those seasons. Figure \ref{fig:dataablation} shows the performance of MTL and STL models when we choose one of eight cultivars and artificially select only a subset of seasons, and train the MultiH and ConcatE model with data from other cultivars not ablated. We also train STL models in the same setup. As expected for STL, with the exception of Cabernet Sauvignon, introducing more seasons of data up to an extent does help in improving performance.\footnote{Note that there is a consistent decrease in performance when going from 20 seasons to ALL. The reasons for this remain to be explored; however, it is likely due to the influence of a small number of unusual seasons.}  

Interestingly we see that an MTL model trained on just 2 or 5 seasons of data for a cultivar outperforms an STL model using all of that cultivar's data. This reflects the fact that MTL is indeed able to leverage the information present in other cultivars to learn a good model for that specific cultivar. In a sense, the data from other cultivars appear to be as valuable as tens of seasons of data for a specific cultivar. We note that the MTL models do not show a consistent improvement in performance across cultivars as the amount of data increases. In some cases, there is even a degradation of performance. This is consistent with the observation that the data from other cultivars seems to have significant value compared to 10s of seasons of a specific cultivars data. 

\subsubsection{MTL vs. Data Aggregation}

Empirically, it has been observed that the performance of deep learning models tends to improve when the training data set size is increased. One could then argue that the performance gains we observe in MTL could be solely attributed to the increase in data set size. We show that this is indeed not the case and that MTL not only captures some common structure between tasks as compared to an STL model presented with all of the cultivars' data aggregated but also takes into account the individual differences of the cultivars. 

\begin{figure}[h!]
    \centering
    \includegraphics[width=1\textwidth]{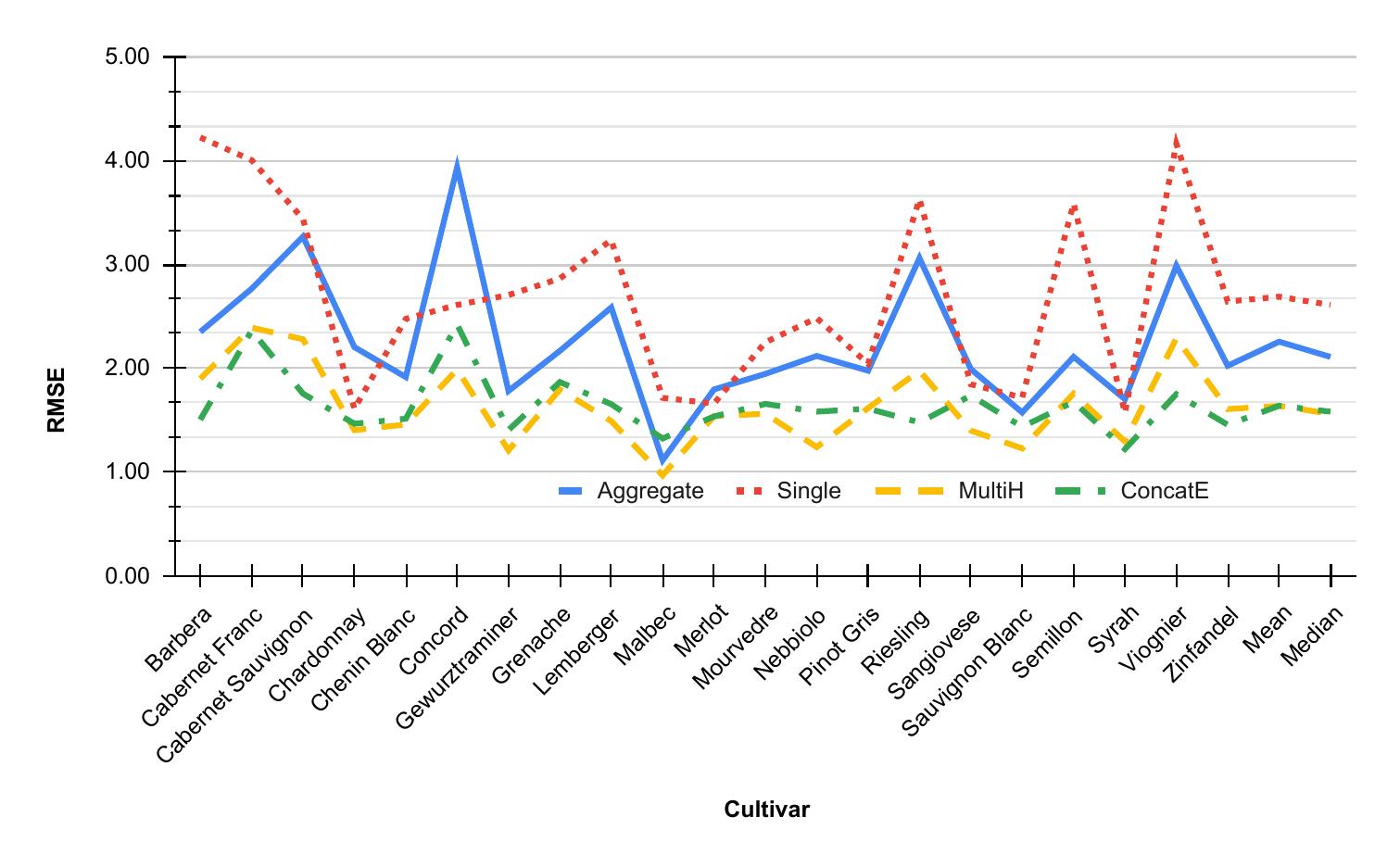} 
    \caption{Comparison of MTL and STL models with another STL model with aggregated data for LTE prediction on grape cultivars.}
    \label{fig:mtlvsdavsstl}
\end{figure}

In Figure \ref{fig:mtlvsdavsstl}, we see that aggregating the data does indeed improve an STL model's performance for the grape LTE prediction task, which validates prior empirical knowledge about deep learning models. We see that both our MultiH and ConcatE models perform better than the data aggregation on average. Similar observations hold for budbreak prediction. 

\subsection{Transfer Learning}
\label{subsec:TL}
We now investigate adaptations of our MTL models to the transfer learning setting (see Section \ref{sec:mtlformulation}). In the context of our application this setting becomes relevant when it is not practical or desirable to continually maintain and aggregate all of the previously encountered datasets. For examples, farmers may be reluctant to share their own data, but willing to run models locally on their own small dataset. In addition, even if maintaining and sharing data were not an issue, the cost of retraining a full MTL model to incorporate a single new data source is significant, especially if training is to be done on a farm managers local machine. This motivates the practical setting of transfer learning, where a previously trained MTL model on source tasks can be combined with a new dataset for a target task to provide predictions specialized to the target.

In what follows, the source tasks correspond to datasets from cultivars that we want to utilize to transfer to a new target cultivar. Transfer learning occurs over two stages. During the transfer preparation stage, the source tasks are used to create a model intended to be used for transfer. During transfer application stage, we are able to use the model produced during transfer preparation and data for the target task to arrive at a new model for the target. Below we describe the adaptations of our MTL models to the transfer setting and then present the experimental results.

For the MultiH model, we use a finetuning strategy for transfer learning, which is perhaps the most common transfer approach used in practice. Finetuning as a technique in deep learning refers to freezing parts or entirety of a model's parameters and optimizing new layers built on top of the network, preserving the useful representations in the frozen model weights. During transfer preparation, MTL is used to learn a multi-head model over the source tasks. During the transfer application stage, that model is augmented with a new prediction head $g_{\theta_{n+1}}$ for the target task initialized with random weights. The data for the target task is then used to train the augmented model end-to-end. The hope is that transfer will be achieved due to the initial shared representation provided by the MTL model learned during transfer preparation. 

For the task embedding model, during transfer preparation we also train a MTL model over the source tasks and denote the learned embedding vector for source cultivar $i$ as $E_i$. Given a new target cultivar at the transfer application stage, we perform transfer by holding the MTL model parameters fixed and identifying a new embedding vector $E_{n+1}$ for the target that optimizes performance on the target's training data. This new embedding vector is then used by the MTL model to make predictions for the target. To keep the new embedding vector within the space of source embedding vectors, we constrain it to be a linear combination of the source embedding vectors. That is, $E_{n+1} = \sum_{i=1}^{n} \alpha_i \cdot E(c_i)$ where the coefficients $\alpha_i$ are initialized randomly. We then optimize the coefficients using gradient descent based on the training loss of the target cultivar.

Finally, we include a baseline transfer learning approach based on model selection. This approach simply trains an STL model for each of the source tasks during transfer preparation. At transfer application, the MSE of each source STL model is evaluated on the target training data and the best STL model is selected to be used for prediction of the target. Such a model selection approach can be expected to perform well when the source cultivars capture the typical variations that may arise in a new target.

Table \ref{tab:tlvsmtlvsstl} presents the results for these transfer approaches along with the corresponding results for standard MTL. We see that transfer via the MultiH model is on par with the corresponding MTL results. Thus, there is little performance loss in not having the source task datasets available along with the target data. However, transfer via the embedding model appears to significantly hurt performance compared to the MTL embedding models and even the STL model, indicating negative transfer. Rather, the baseline model-selection method provides non-trivial transfer performance overall. It generally outperforms STL, while still being outperformed by the MTL methods and MultiH transfer. We do not yet have a strong explanation for the poor performance of transfer with the embedding model. However, the non-trivial performance of model selection indicates that the issue may be with the optimization approach for the target embedding. In particular, model selection corresponds to a specific setting of the transfer coefficients (zeros assigned to all but one of the source models). However, the optimization of the coefficients settled on a much lower performing setting.

\begin{table}[htp]
\centering
\resizebox{1.0\columnwidth}{!}{
\begin{tabular}{|l|l|l|l|l|l|l|}
\hline
Cultivar             & ConcatE\_transfer & MultiH\_Transfer & Model   Selection & ConcatE\_MTL  & MultiH\_MTL   & Single  \\ 
\hline
Barbera              & 2.52              & 1.89             & 1.66              & \textbf{1.50} & 1.90          & 4.22    \\ 
\hline
Cabernet   Franc     & 3.86              & 2.44             & \textbf{2.10}     & 2.36          & 2.39          & 4.00    \\ 
\hline
Cabernet   Sauvignon & 2.78              & 2.29             & \textbf{1.27}     & 1.76          & 2.28          & 3.44    \\ 
\hline
Chardonnay           & 2.61              & \textbf{1.29}    & 2.71              & 1.47          & 1.40          & 1.61    \\ 
\hline
Chenin   Blanc       & 2.26              & 1.51             & 2.15              & 1.51          & \textbf{1.46} & 2.48    \\ 
\hline
Concord              & 5.80              & 2.23             & 3.66              & 2.42          & \textbf{1.99} & 2.61    \\ 
\hline
Gewurztraminer       & 2.99              & 1.46             & 2.53              & 1.41          & \textbf{1.21} & 2.71    \\ 
\hline
Grenache             & 2.18              & 1.80             & 1.84              & 1.87          & \textbf{1.79} & 2.86    \\ 
\hline
Lemberger            & 4.18              & 1.65             & 2.60              & 1.65          & \textbf{1.49} & 3.24    \\ 
\hline
Malbec               & 4.10              & 1.04             & 1.18              & 1.32          & \textbf{0.97} & 1.71    \\ 
\hline
Merlot               & 2.38              & \textbf{1.43}    & 1.48              & 1.54          & 1.54          & 1.66    \\ 
\hline
Mourvedre            & 2.94              & 1.63             & 1.95              & 1.66          & \textbf{1.56} & 2.25    \\ 
\hline
Nebbiolo             & 4.09              & 1.65             & 1.86              & 1.58          & \textbf{1.24} & 2.48    \\ 
\hline
Pinot   Gris         & 2.75              & \textbf{1.54}    & 1.67              & 1.61          & 1.62          & 2.04    \\ 
\hline
Riesling             & 3.25              & 1.66             & 2.68              & \textbf{1.48} & 1.98          & 3.63    \\ 
\hline
Sangiovese           & 2.70              & \textbf{1.35}    & 1.89              & 1.74          & 1.40          & 1.85    \\ 
\hline
Sauvignon   Blanc    & 2.66              & 1.25             & 2.21              & 1.44          & \textbf{1.23} & 1.72    \\ 
\hline
Semillon             & 3.13              & \textbf{1.40}    & 2.24              & 1.68          & 1.76          & 3.59    \\ 
\hline
Syrah                & 2.19              & 1.29             & 1.57              & \textbf{1.22} & 1.29          & 1.57    \\ 
\hline
Viognier             & 3.00              & 1.77             & 1.95              & \textbf{1.75} & 2.29          & 4.16    \\ 
\hline
Zinfandel            & 3.70              & 1.79             & 1.69              & \textbf{1.46} & 1.61          & 2.64    \\ 
\hline
Mean                 & 3.15              & \textbf{1.64}    & 2.04              & \textbf{1.64} & \textbf{1.64} & 2.69    \\ 
\hline
Median               & 2.94              & 1.63             & 1.95              & 1.58          & \textbf{1.56} & 2.61    \\
\hline
\end{tabular}}
\caption{Comparing Transfer Learning approaches to STL and MTL, the metric is Root Mean Squared Error (Lower is Better)}
\label{tab:tlvsmtlvsstl}
\end{table}

\section{Deployment and Testing}
\label{sec:deployment}
Growers and farmers use AgWeatherNet \cite{AgWeatherNet} and WSU Viticulture and Enology \cite{viticulture_wsu} websites to monitor cold hardiness through the deployment of the Ferguson model and publication of real LTE values, respectively. We have finalized our MTL models proposed in this paper and deployed them onto AgWeatherNet for the 2022-2023 season for beta testing. A subset of users were given access to the models and asked for feedback at the end of the season. Farmers were able to use weather data from the nearest AgWeatherNet station.

\section{Summary and Initial Deployment}
\label{sec:conclusions}
We showed that multi-task learning using straightforward wrapper-style approaches is an effective approach for cold hardiness and budbreak prediction for grape cultivars. In particular, our MTL models consistently outperform STL models and the state-of-the-art scientific model without relying on expert domain knowledge. In particular, we found that the MTL models are able to outperform STL models with only a tiny fraction of the training data for a given test cultivar. We also observed that the best performing MTL model varied across the different prediction problems considered (Grape LTE, Grape Budbreak). This indicates that it is worth spending some validation effort, when possible, to select the best MTL model for a new application. We also observed that a straightforward fine-tuning approch for transfer using the multi-head model was quite effective, while our initial efforts at transfer via the embedding model were not effective. 

From an application perspective, the most important future work is to widely deploy these models for use by farmers. We have already conducted initial beta testing for a selected set of users on the AgWeatherNet \cite{AgWeatherNet} weather network during the 2022-2023 season. AgWeatherNet is used as a primary information source for many farmers in the Pacific Northwest and has previously only included the Ferguson model for grape cold-hardiness prediction. Our beta testing allowed the selected farmers to utilize the Ferguson and our MTL models and provide feedback on their assessment of the predictions. Those results are currently being analyzed to identify the most important directions for improvement to the interfaces and models themselves.

\backmatter

\noindent 
\bigskip

\newpage
 
\begin{appendices}

\section{Comparison of MTL Embedding Models}
\label{sec:lab-comp}

Due to computational constraints, we were unable to consider all the MTL Task embedding variant models for our planned suite of experiments. Table \ref{tab:compareTE} refers to the performance of the MTL Embedding Models for the grape LTE prediction problem. On average, we notice that the ConcatE variant model performs better than other variants. We also see that the multiplicative embedding comes close and could be considered in future work as a viable alternative to the ConcatE model. 

\begin{longtable}{|l|c|c|c|c|c|}
\hline \multicolumn{1}{|c|}{\textbf{Cultivar}} & \multicolumn{1}{c|}{\textbf{MultE}} & \multicolumn{1}{c|}{\textbf{ConcatE}} & \multicolumn{1}{c|}{\textbf{AddE}} \\ \hline 

Barbera            & 1.93 & 1.50 & 2.08 \\ \hline
Cabernet Franc     & 4.85 & 2.36 & 3.50 \\ \hline
Cabernet Sauvignon & 2.93 & 1.75 & 1.82 \\ \hline
Chardonnay         & 1.33 & 1.47 & 1.45 \\ \hline
Chenin Blanc       & 1.85 & 1.51 & 1.57 \\ \hline
Concord            & 2.33 & 2.42 & 2.33 \\ \hline
Gewurztraminer     & 1.97 & 1.41 & 1.67 \\ \hline
Grenache           & 3.07 & 1.87 & 2.18 \\ \hline
Lemberger          & 3.01 & 1.65 & 2.25 \\ \hline
Malbec             & 1.81 & 1.32 & 1.32 \\ \hline
Merlot             & 1.75 & 1.54 & 1.40 \\ \hline
Mourvedre          & 1.85 & 1.66 & 1.70 \\ \hline
Nebbiolo           & 2.37 & 1.58 & 1.87 \\ \hline
Pinot Gris         & 2.08 & 1.61 & 1.63 \\ \hline
Riesling           & 2.80 & 1.48 & 1.77 \\ \hline
Sangiovese         & 1.66 & 1.74 & 1.71 \\ \hline
Sauvignon Blanc    & 1.33 & 1.44 & 1.53 \\ \hline
Semillon           & 2.38 & 1.68 & 1.42 \\ \hline
Syrah              & 1.23 & 1.22 & 1.28 \\ \hline
Viognier           & 3.90 & 1.75 & 2.30 \\ \hline
Zinfandel          & 3.10 & 1.45 & 1.57 \\ \hline
Mean               & 2.19 & 1.83 & 1.92 \\ \hline
Median             & 1.92 & 1.70 & 1.82 \\ \hline
\caption{Comparison of different Task Embedding MTL models on grape cultivars for LTE prediction. Note that RMSE is the metric.}
\label{tab:compareTE}
\end{longtable}

\section{Impact of Task Data Set Size - Full Results}
\label{sec:full-dataset-size}

\begin{longtable}{@{}|l|l|l|l|l|l|@{}}
\hline
\thead{\textbf{Cultivar}} & \thead{2} & \thead{5} & \thead{10} & \thead{20} & \thead{All} \\ \hline
Barbera\_Single                                             & 6.23                       & 5.57                       & 4.91                        & 4.22                        & 4.22                         \\ \hline
Barbera\_ConcatE                                            & 2.05                       & 1.94                       & 1.84                        & 1.50                        & 1.50                         \\ \hline
Barbera\_MultiH                                             & 2.04                       & 1.96                       & 2.03                        & 1.90                        & 1.90                         \\ \hline
Cabernet Franc\_Single                                      & 4.01                       & 4.01                       & 4.01                        & 4.01                        & 4.01                         \\ \hline
Cabernet Franc\_ConcatE                                     & 2.36                       & 2.36                       & 2.36                        & 2.36                        & 2.36                         \\ \hline
Cabernet Franc\_MultiH                                      & 2.39                       & 2.39                       & 2.39                        & 2.39                        & 2.39                         \\ \hline
Cabernet Sauvignon\_Single                                  & 2.50                       & 2.23                       & 3.70                        & 3.50                        & 3.44                         \\ \hline
Cabernet Sauvignon\_ConcatE                                 & 2.46                       & 1.81                       & 2.02                        & 1.72                        & 1.76                         \\ \hline
Cabernet Sauvignon\_MultiH                                  & 2.26                       & 2.29                       & 2.39                        & 2.49                        & 2.28                         \\ \hline
Chardonnay\_Single                                          & 2.30                       & 1.88                       & 1.84                        & 1.66                        & 1.61                         \\ \hline
Chardonnay\_ConcatE                                         & 1.62                       & 1.44                       & 1.52                        & 1.42                        & 1.47                         \\ \hline
Chardonnay\_MultiH                                          & 1.20                       & 1.53                       & 1.50                        & 1.34                        & 1.40                         \\ \hline
Chenin Blanc\_Single                                        & 3.53                       & 3.49                       & 2.59                        & 2.48                        & 2.48                         \\ \hline
Chenin Blanc\_ConcatE                                       & 1.49                       & 1.85                       & 1.27                        & 1.51                        & 1.51                         \\ \hline
Chenin Blanc\_MultiH                                        & 1.65                       & 1.55                       & 1.62                        & 1.46                        & 1.46                         \\ \hline
Concord\_Single                                             & 4.02                       & 3.53                       & 2.72                        & 2.51                        & 2.61                         \\ \hline
Concord\_ConcatE                                            & 3.16                       & 2.56                       & 2.40                        & 2.27                        & 2.42                         \\ \hline
Concord\_MultiH                                             & 2.24                       & 2.33                       & 2.06                        & 2.00                        & 1.99                         \\ \hline
Gewurztraminer\_Single                                      & 3.36                       & 2.68                       & 2.71                        & 2.71                        & 2.71                         \\ \hline
Gewurztraminer\_ConcatE                                     & 1.51                       & 1.45                       & 1.41                        & 1.41                        & 1.41                         \\ \hline
Gewurztraminer\_MultiH                                      & 1.35                       & 1.43                       & 1.21                        & 1.21                        & 1.21                         \\ \hline
Grenache\_Single                                            & 3.17                       & 3.54                       & 2.30                        & 2.86                        & 2.86                         \\ \hline
Grenache\_ConcatE                                           & 1.77                       & 1.85                       & 1.97                        & 1.87                        & 1.87                         \\ \hline
Grenache\_MultiH                                            & 2.26                       & 1.92                       & 1.80                        & 1.79                        & 1.79                         \\ \hline
Lemberger\_Single                                           & 5.08                       & 3.24                       & 3.24                        & 3.24                        & 3.24                         \\ \hline
Lemberger\_ConcatE                                          & 1.85                       & 1.65                       & 1.65                        & 1.65                        & 1.65                         \\ \hline
Lemberger\_MultiH                                           & 2.69                       & 1.49                       & 1.49                        & 1.49                        & 1.49                         \\ \hline
Malbec\_Single                                              & 2.40                       & 2.01                       & 1.56                        & 1.72                        & 1.72                         \\ \hline
Malbec\_ConcatE                                             & 1.50                       & 1.49                       & 1.30                        & 1.32                        & 1.32                         \\ \hline
Malbec\_MultiH                                              & 1.24                       & 1.12                       & 1.03                        & 0.97                        & 0.97                         \\ \hline
Merlot\_Single                                              & 2.33                       & 1.72                       & 1.74                        & 1.69                        & 1.66                         \\ \hline
Merlot\_ConcatE                                             & 1.73                       & 1.51                       & 1.54                        & 1.35                        & 1.54                         \\ \hline
Merlot\_MultiH                                              & 1.67                       & 1.70                       & 1.44                        & 1.43                        & 1.54                         \\ \hline
Mourvedre\_Single                                           & 2.33                       & 2.81                       & 2.25                        & 2.25                        & 2.25                         \\ \hline
Mourvedre\_ConcatE                                          & 2.21                       & 1.68                       & 1.66                        & 1.66                        & 1.66                         \\ \hline
Mourvedre\_MultiH                                           & 1.92                       & 1.61                       & 1.56                        & 1.56                        & 1.56                         \\ \hline
Nebbiolo\_Single                                            & 4.44                       & 3.64                       & 2.67                        & 2.48                        & 2.48                         \\ \hline
Nebbiolo\_ConcatE                                           & 1.80                       & 1.53                       & 1.74                        & 1.58                        & 1.58                         \\ \hline
Nebbiolo\_MultiH                                            & 1.65                       & 1.37                       & 1.26                        & 1.24                        & 1.24                         \\ \hline
Pinot Gris\_Single                                          & 3.35                       & 2.66                       & 2.27                        & 2.04                        & 2.04                         \\ \hline
Pinot Gris\_ConcatE                                         & 1.56                       & 1.53                       & 1.59                        & 1.61                        & 1.61                         \\ \hline
Pinot Gris\_MultiH                                          & 1.72                       & 1.62                       & 1.54                        & 1.62                        & 1.62                         \\ \hline
Riesling\_Single                                            & 5.30                       & 4.25                       & 3.50                        & 3.10                        & 3.63                         \\ \hline
Riesling\_ConcatE                                           & 2.15                       & 2.51                       & 2.00                        & 1.92                        & 1.48                         \\ \hline
Riesling\_MultiH                                            & 3.25                       & 1.97                       & 2.16                        & 1.77                        & 1.98                         \\ \hline
Sangiovese\_Single                                          & 2.48                       & 2.00                       & 1.62                        & 1.85                        & 1.85                         \\ \hline
Sangiovese\_ConcatE                                         & 1.52                       & 1.29                       & 1.36                        & 1.74                        & 1.74                         \\ \hline
Sangiovese\_MultiH                                          & 1.18                       & 1.33                       & 1.69                        & 1.40                        & 1.40                         \\ \hline
Sauvignon Blanc\_Single                                     & 2.49                       & 2.23                       & 1.72                        & 1.72                        & 1.72                         \\ \hline
Sauvignon Blanc\_ConcatE                                    & 1.72                       & 1.45                       & 1.44                        & 1.44                        & 1.44                         \\ \hline
Sauvignon Blanc\_MultiH                                     & 1.29                       & 1.10                       & 1.23                        & 1.23                        & 1.23                         \\ \hline
Semillon\_Single                                            & 5.06                       & 3.07                       & 4.02                        & 3.59                        & 3.59                         \\ \hline
Semillon\_ConcatE                                           & 1.64                       & 1.79                       & 1.53                        & 1.68                        & 1.68                         \\ \hline
Semillon\_MultiH                                            & 1.95                       & 1.80                       & 1.56                        & 1.76                        & 1.76                         \\ \hline
Syrah\_Single                                               & 2.26                       & 2.14                       & 1.74                        & 1.57                        & 1.57                         \\ \hline
Syrah\_ConcatE                                              & 1.44                       & 1.21                       & 1.19                        & 1.22                        & 1.22                         \\ \hline
Syrah\_MultiH                                               & 2.44                       & 1.37                       & 1.21                        & 1.29                        & 1.29                         \\ \hline
Viognier\_Single                                            & 4.23                       & 4.69                       & 4.97                        & 4.16                        & 4.16                         \\ \hline
Viognier\_ConcatE                                           & 2.21                       & 1.63                       & 1.92                        & 1.75                        & 1.75                         \\ \hline
Viognier\_MultiH                                            & 2.29                       & 2.73                       & 3.00                        & 2.29                        & 2.29                         \\ \hline
Zinfandel\_Single                                           & 3.17                       & 4.13                       & 2.66                        & 2.64                        & 2.64                         \\ \hline
Zinfandel\_ConcatE                                          & 1.86                       & 1.62                       & 1.52                        & 1.46                        & 1.46                         \\ \hline
Zinfandel\_MultiH                                           & 1.55                       & 1.38                       & 1.40                        & 1.61                        & 1.61                         \\ \hline
\caption{Full results for dataset size ablation experiments for STL and MTL on all Grape cultivars for LTE Prediction.}
\label{tab:myfirstlongtable}
\end{longtable}
Table \ref{tab:myfirstlongtable} refers to the dataset size ablation experiment conducted for all 21 grape cultivars. A cursory glance over the table shows that an MTL model trained with just 2/5 seasons of a cultivar can surpass the performance of an STL model which uses all the seasons of that particular cultivar. Un-intuitively, we see a degradation in performance when the number of seasons increases from 20 to all of the seasons. We posit that this might be due to the presence of noisy weather information for certain seasons that were not present in the chosen 20 seasons. 

\end{appendices}

\bibliography{sn-bibliography}

\end{document}